\documentclass[10pt,letterpaper]{article}
\usepackage[top=0.85in,left=2.75in,footskip=0.75in]{geometry}

\usepackage{amsmath,amssymb}

\usepackage{changepage}

\usepackage{textcomp,marvosym}

\usepackage{cite}

\usepackage{nameref,hyperref}

\usepackage[right]{lineno}

\usepackage[nopatch=eqnum]{microtype}
\DisableLigatures[f]{encoding = *, family = * }

\usepackage[table]{xcolor}

\usepackage{booktabs} 
\usepackage{ragged2e} 
\usepackage{float}

\usepackage{easyReview}

\usepackage{array}

\newcolumntype{+}{!{\vrule width 2pt}}

\newlength\savedwidth

\raggedright
\usepackage[aboveskip=1pt,labelfont=bf,labelsep=period,justification=raggedright,singlelinecheck=off]{caption}

\makeatletter
\renewcommand{\@biblabel}[1]{\quad#1.}
\makeatother

\usepackage{lastpage,fancyhdr,graphicx}
\usepackage{epstopdf}
\fancyheadoffset[L]{2.25in}
\fancyfootoffset[L]{2.25in}
\begin{document}
\vspace*{0.2in}

\begin{flushleft}
{\Large
\textbf\newline{Deep Learning for Early Alzheimer Disease Detection with MRI Scans} 
}
\newline
\\
Mohammad Rafsan\textsuperscript{1\Yinyang},
Tamer Oraby\textsuperscript{1\Yinyang},
Upal Roy\textsuperscript{1\textpilcrow},
Sanjeev Kumar\textsuperscript{1\textpilcrow}
Hansapani Rodrigo\textsuperscript{1\Yinyang},

\bigskip
\textbf{1} University of Texas Rio Grande Valley, Edinburg, Texas, United States of America.
\\
\bigskip

%
%
\Yinyang These authors contributed equally to this work.




\textpilcrow Methodology, Interpretation, Final manuscript review.

* mohammad.rafsan01@utrgv.edu\\
* tamer.oraby@utrgv.edu\\
* upal.roy@utrgv.edu\\
* sj.kumar@utrgv.edu\\
* hansapani.rodrigo@utrgv.edu\\


%

\end{flushleft}
\section*{Abstract}
\justifying
Alzheimer's Disease is a neurodegenerative condition characterized by dementia and impairment in neurological function. The study primarily focuses on the individuals above age 40, affecting their memory, behavior, and cognitive processes of the brain. Alzheimer's disease requires diagnosis by a detailed assessment of MRI scans and neuropsychological tests of the patients. This project compares existing deep learning models in the pursuit of enhancing the accuracy and efficiency of AD diagnosis, specifically focusing on the Convolutional Neural Network, Bayesian Convolutional Neural Network, and the U-net model with the Open Access Series of Imaging Studies brain MRI dataset. Besides, to ensure robustness and reliability in the model evaluations, we address the challenge of imbalance in data. We then perform rigorous evaluation to determine strengths and weaknesses for each model by considering sensitivity, specificity, and computational efficiency. This comparative analysis would shed light on the future role of AI in revolutionizing AD diagnostics but also paved ways for future innovation in medical imaging and the management of neurodegenerative diseases.

\section*{Author summary}

\subsection*{Why was this study done?}
\justifying
\begin{itemize}
    \item Alzheimer's disease is the most common cause of dementia, affecting millions worldwide, and yet the diagnosis remains challenging in the early stage.
    \item The state-of-the-art methods in diagnosis include MRI scans and neuropsychological tests that require significant expertise and time.
    \item The current study focuses on deep learning models and their possible application to improving the accuracy of early-stage Alzheimer's disease diagnosis.
\end{itemize}

\subsection*{What did the researchers do and find?}
\begin{itemize}
    \item We applied three deep learning models, namely Convolutional Neural Networks, Bayesian Convolutional Neural Network, and U-Net on MRI scans from the OASIS brain MRI dataset.
    \item The proposed models are trained on the balanced dataset using SMOTE-Tomek and then tested on accuracy, precision, recall, and F1-score.
    \item Bayesian Convolutional Neural Network achieved an accuracy above 95\%, while CNN and U-Net scored the next highest position.
    \item We used Gradient-weighted Class Activation Mapping to help us visualize those areas of the brain which contributed most to the model's predictions, hence helping in the interpretability of the deep learning models concerning their diagnostic relevance.

\end{itemize}

\subsection*{What do these findings mean?}
\begin{itemize}
    \item The findings demonstrate the promise of Artificial Intelligence, particularly Bayesian Convolutional Neural Network, in improving the early diagnosis of Alzheimer’s Disease.
    \item The study emphasizes the potential for Artificial Intelligence-driven approaches to advance clinical diagnostics, especially for neurodegenerative diseases.
    \item Grad-CAM provides a visual explanation of the model's prediction by highlighting the most relevant regions in the brain MRI images. The capability of visually interpreting the model's focused areas by Grad-CAM could be an important aid to assist in early diagnosis and, hence, interventions may be timely and appropriate.
    \item Future work will involve developing, in collaboration with oncologists, a masking method for brain MRI to improve the focus on critical brain regions, and identifying key variables for early prediction of AD, allowing more targeted and hence effective diagnostic strategies that have practical applications in healthcare.
\end{itemize}



\section*{Introduction}
\justifying
Alzheimer's disease is a progressive neurodegenerative disorder and the most common cause of dementia among older adults, affecting memory, thinking, and behavior. According to the World Health Organization, AD accounts for 60-70\% of dementia cases worldwide, making it a significant public health concern~\cite{WHO2020}. Early diagnosis and monitoring of AD are important in managing disease progression and planning appropriate treatment strategies. The diagnosis of AD has conventionally been made by clinical evaluations, including neuropsychological testing and neuroimaging techniques such as MRI scans. While these methods are effective, they demand a great deal of expertise and time for analysis. AD is one of the most important challenges in the field of neurodegenerative diseases, with continuous disturbance in cognitive decline and loss of neurons. As the most common cause of dementia, AD significantly affects not only the quality of life of those concerned but also burdens health-care systems and caregivers all over the world. For this reason, the establishment of more effective therapies depends essentially on early detection to properly manage its course and attenuate its impacts.

In the last few years, AI, especially Deep Learning, has emerged as a force of transformation in many fields, including healthcare and medical imaging. These DL models, known for their capability of learning complex patterns from big datasets, have shown great promise in enhancing the analysis of medical images, including those used in the diagnosis of AD. Among the numerous DL architectures, CNNs, BayesianCNNs and U-net models have demonstrated great performance in medical image analysis. BayesianCNNs add a probabilistic method using BayesCNN-an extension of Bayes by Backprop~\cite{blundell2015weight}-which provides probabilities of the weights in neural networks and allows for uncertainty quantification in the predicted output, which is one of the most useful applications in medical diagnosis where making decisions under uncertainty is always present~\cite{graves2011practical}. The U-net architecture, initially designed for biomedical image segmentation, showed efficiency in the analysis of medical images at the level of detailed investigation, including those tasks related to the identification of neurodegenerative diseases markers~\cite{ronneberger2015u}.

The focus of this project is the comparison of these sophisticated DL models to improve the diagnosis and monitoring of AD. Powered by an open-access series of imaging studies, the brain MRI dataset will include subjects with all stages of AD~\cite{marcus2010open}, the research will investigate whether BayesianCNN has advantages over a U-net model in predicting the future state of the disease or whether there are no statistical differences in results. The OASIS dataset, for example, is an excellent foundation to build upon in establishing how well these DL models could work in the wild due to its comprehensive neuroimaging data collection along with clinical assessments. Thus, through a comparative analysis, the research work will contribute to an update in the literature regarding AI-driven diagnostics in neurodegenerative diseases while also exploring the capability of DL in bringing about a revolution in early detection and treatment planning for AD. This project uses the publicly available OASIS brain MRI dataset, which includes MRI scans of subjects with varying stages of AD. Comparing the performance of the CNN, BayesianCNN, and U-net models on this dataset, this study has aimed to evaluate their potentials in supporting early detection and monitoring of AD, hence contributing to the development of AI-driven diagnostic tools in neurology.

The paper has a smooth flow, from the broad literature review to the very foundation of the study itself. It follows the Materials and Methods section, detailing the approach and techniques involved in the study. Next, the paper presents results for the different models with findings from experiments. Finally, the discussion and conclusion section integrates the insights from the results, summarizing the contributions and implications of the research, and pointing out any possible future work.

\section*{Literature Review}
\justifying
Accurate medical picture categorization is a challenging undertaking due to the intricate process of acquiring medical data sets~\cite{alzubaidi2020towards}. Medical data sets, in contrast to other types of data sets, are created by qualified professionals and include confidential and sensitive patient information that is not allowed to be made public. Because of this, organizations and institutions that provide medical data sets, such as the Alzheimer's Disease Neuroimaging Initiative (ADNI)~\cite{weiner2013alzheimer} and the OASIS ~\cite{marcus2010open}, have screening procedures for accessing their data sets. These procedures require the researcher to fill out an application and agree to terms, which restricts the researcher's ability to use the data only for research purposes~\cite{shereen2021proposed}, ~\cite{mohammed2021multi}. Since it is difficult to assemble a data set with an equal amount of participants with health and illness samples, medical data sets are intrinsically extremely unbalanced. The methods for solving this issue are somewhat difficult by themselves~\cite{battineni2021deep}. 

Eskildsen et al.~\cite{eskildsen2013prediction} employed cortical thickness measurements to determine distinct patterns of atrophy, and attributes were picked from these patterns to predict AD in patients with moderate cognitive impairment (MCI). A deep learning-based technique for evaluating the Mini-Mental State Examination (MMSE) using resting-state functional magnetic resonance imaging (rsfMRI) was developed by Duc et al~\cite{duc20203d}. The system produced positive findings for the diagnosis of AD. Islam et al. ~\cite{islam2017novel} used automatic identification and categorization of Alzheimer's illness using a deep CNN model. Its model draws inspiration from the Inception-V4 network~\cite{szegedy2017inception}. Using brain magnetic resonance imaging (MRI), Taher et al. ~\cite{ghazal2022alzheimer} suggested method for Alzheimer disease diagnosis utilizes transfer learning to multi-class classification. They have modified the CNN based pre-trained AlexNet network ~\cite{alom2018history} consisting of an eight-layer network with learnable parameters, consisting of three fully connected layers and five convolutional layers that combine max pooling. Sarraf et al. ~\cite{sarraf2016classification} detected AD using the ADNI dataset using fMRI data and a deep LeNet~\cite{albawi2017understanding} model. Ruhul et al. ~\cite{hazarika2021improved} have modified the traditional LeNet network. They have developed a distinct layer to carry out the Min-Pooling function. The layers of MinPooling and MaxPooling are then joined together and added concatenated layers to LeNet in place of all MaxPooling Layers. The use of several approximation techniques for the intractable true posterior probability distribution \(p(w|D)\) has been investigated in the past when applying Bayesian approaches to neural networks. A number of maximum-a-posteriori (MAP) algorithms for neural networks were first proposed by ~\cite{buntine1991bayesian} . In order to promote smoothness of the resultant approximate posterior probability distribution, they were also the first to propose second order derivatives in the prior probability distribution \(p(w)\).

MRI imaging plays a pivotal role in differentiating between normal and Alzheimer's affected brains by providing detailed insights into structural changes. In healthy brains, MRI scans typically exhibit well-maintained cortical structures and hippocampal volumes, with no signs of significant atrophy. In contrast, AD is often characterized by marked cortical atrophy, particularly in the temporal and parietal lobes, which correlates with cognitive decline~\cite{radiopaedia_alz}. Hippocampal atrophy is one of the earliest and most prominent markers of AD, making it a crucial biomarker for diagnosis~\cite{radiology_assistant}. Additionally, enlarged ventricles are commonly observed due to the loss of brain tissue, further indicating neurodegeneration~\cite{alz_research_uk}. Clinicians interpret MRI scans by visually assessing patterns of atrophy, quantitatively measuring brain structure volumes, and comparing changes over time to track disease progression. Such assessments are often integrated with cognitive tests to provide a comprehensive evaluation for diagnosing Alzheimer's~\cite{alz_research_uk, radiology_assistant}.


In general, this research will contribute to the adaptation and optimization of deep learning models for neuroimaging and the development of methodologies to handle class imbalance in medical imaging. We will investigate different deep learning models to show their potential for the identification of early AD biomarkers, hence showing the flexibility and wide applicability of the DL technologies in medical diagnostics. This includes reconfiguration and optimization of the architecture and parameters of well-established CNNs and U-net architectures for applications in neuroimaging to achieve the best performance. Furthermore, several techniques are developed and implemented to better cope with biased data distribution in datasets of medical images, extending our contributions toward model development for data processing and training effectiveness.

More precisely, this work compares three competing deep learning models that are currently popular in computer vision: ADD-Net based on CNN, BayesianCNN, and U-net. To the best of our knowledge, our study explored, for the first time, the use of BayesianCNN and U-net for their capability to be used in the detection of early AD. Our analysis incorporates the use of ADD-Net along with a hybrid SMOTE-Tomek approach on the OASIS dataset. We also introduce SMOTE-Tomek for BayesianCNN to conclude between BayesianCNN and the U-Net model as tools to effectively detect early AD. These models are implemented in Python 3.8.10, TensorFlow 2.17.0, Keras 3.4.1, and Scikit-learn 2.2.0, among other libraries. Standard metrics for performance evaluation in classification tasks include Accuracy, F1-score, Recall, and Precision. These will allow us to compare different models' performances in effectively detecting early Alzheimer's Disease using the OASIS brain MRI dataset.
We will also describe how such models improve with hyperparameter fine-tuning, optimization of the training processes, and new data-handling techniques that might promise more valid and reliable early detection of AD.

\vspace{0.5cm}

The source code of this paper is available on \textbf{Github}:~
\href{https://github.com/RafusaN/DL-Alzheimer.git}{DL-Alzheimer}

\section*{Materials and methods}
\justifying
In this section are the details of the methodologies adopted in this research deal with the effectiveness of DL techniques in early detection through the use of neuroimaging data. Early markers of AD in imaging data are complex and require a strong, advanced model to make reliable diagnostic predictions.

\subsection*{Data Description}
\justifying
The present work is based on data from the Open Access Series of Imaging Studies
OASIS~\cite{oasis_brains}, a publicly accessible repository with MRI data from nondemented and from subjects with different stages of AD. In summary, the dataset comprises neuroimaging carried out on 1378 subjects (ages 18 - 96, median = 54 years, IQR: 51 years), separated according to the Clinical Dementia Rating (CDR) that varies from non-demented subjects to moderate AD cases. For each subject, various images are available; among them, one can find an average image, that is, a motion-corrected coregistered average of all available data. From the OASIS raw data, the original \texttt{.img} and \texttt{.hdr} files were converted into Nifti format \texttt{.nii} by the free, non-commercial FSL (FMRIB Software Library). The MRI scans in \texttt{.nii} format were then converted into \texttt{.jpg} files using NiBabel (Python), a library of Python programming language designed to make the work of reading and manipulating Nifti format easy. Using NiBabel, one may extract slices of data and save them in image formats such as \texttt{.jpg}.

Clinical diagnosis was made according to the CDR scale and expressed in terms
without resorting to psychometric tests and excluding other causes of dementia \cite{morris1993clinical}. Diagnosis of AD required evidence of progressive loss of memory and decline in other cognitive functions. CDR provides scores for memory, orientation, judgment, community affairs, home and hobbies, and personal care. These scores then give a global CDR: 0 represents no dementia, and the scores 0.5, 1, 2 and 3 correspond to very mild, mild, moderate and severe dementia, respectively~\cite{marcus2010open}. Given the abovementioned CDR scores, the study identifies four classes of dementia, which are respectively, NOD, VMD, MD and MOD, giving rise to a total of 11655 images to be considered for the study. The distribution of the 11,655 images across categories of dementia is given below: ‘Mild Dementia’ comprising 1,573 images, ‘Moderate Dementia’ consists of 124 images, ‘Non-Demented’ comprises 5,849 images, and ‘Very Mild Dementia’ accounts for 4,109 images.

\subsection*{Deep Neural Networks (DNNs)}
\justifying
Deep neural networks are a class of machine learning algorithms that learn to perform tasks by learning from examples in a way inspired by the thinking process of a human mind. Any feed-forward network with a number of hidden layers is called DNN.
DNNs would simply mean multiple layers of artificial neurons or nodes, which are math functions emulating the neural activities of the human brain. Each additional layer increases the feature extraction and processing on the input data, with early layers identifying simple features and deeper layers recognizing complex patterns; thus, it learns from a wide array of data and makes decisions.

\begin{figure}[h]
    \centering
    \includegraphics[scale=0.7]{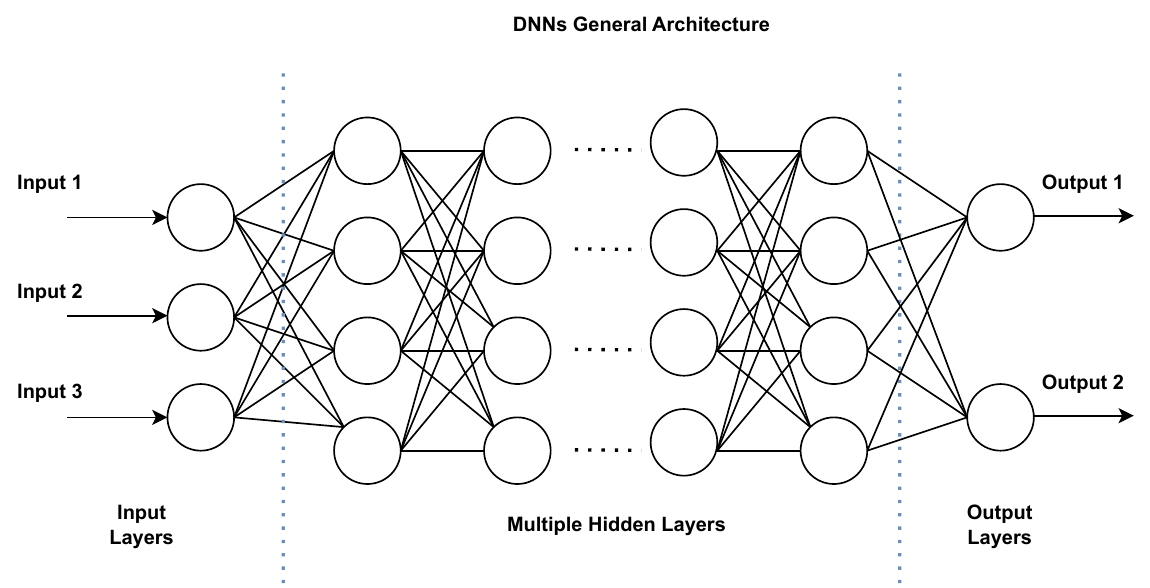} 
    \captionsetup{justification=centering}
    \caption[DNN Model Architecture]{\bf General model architecture of a DNN}{The figure illustrates a typical DNN with input layers, multiple hidden layers, and output layers, showing the interconnected structure used for learning complex patterns in data.}%
    \label{fig:DNN}
\end{figure}

The architecture of a Deep Neural Network (DNN) can be mathematically represented as follows:

\begin{enumerate}
    \item \textbf{Input Layer:} Let the input vector be denoted as $\mathbf{x} \in \mathbb{R}^{d}$, where $d$ is the dimension of the input, which serves as the input to the first layer.
    
    \item \textbf{Hidden Layers:} The network consists of $L-1$ hidden layers, where $L$ is the total number of layers. Each hidden layer $l \in \{1, 2, \ldots, L-1\}$ performs two main operations: a linear transformation followed by a non-linear activation. The operations for each hidden layer $l$ are detailed below:
    
    \begin{itemize}
        \item \textbf{Linear Transformation:} The output from the previous layer $\mathbf{a}^{(l-1)}$ (or $\mathbf{x}$ for $l=1$) is transformed linearly using a weight matrix $\mathbf{W}^{(l)} \in \mathbb{R}^{n_l \times n_{l-1}}$ and a bias vector $\mathbf{b}^{(l)} \in \mathbb{R}^{n_l}$, where $n_l$ and $n_{l-1}$ are the number of neurons in layers $l$ and $l-1$, respectively. The linear transformation is represented as:
        
        \begin{equation}
            \mathbf{z}^{(l)} = \mathbf{W}^{(l)} \mathbf{a}^{(l-1)} + \mathbf{b}^{(l)}
        \end{equation}
        
        \item \textbf{Activation:} The result of the linear transformation $\mathbf{z}^{(l)}$ is then passed through a non-linear activation function $\sigma^{(l)}$ to produce the output of layer $l$:
        
        \begin{equation}
            \mathbf{a}^{(l)} = \sigma^{(l)}(\mathbf{z}^{(l)})
        \end{equation}
    \end{itemize}
    
    \item \textbf{Output Layer:} The final layer $L$ uses an activation function $\sigma^{(L)}$ suitable for the specific task to output the network's prediction, denoted as $\hat{\mathbf{y}}$. This output is computed as follows:
    
    \begin{equation}
        \hat{\mathbf{y}} = \sigma^{(L)}(\mathbf{z}^{(L)}) = \sigma^{(L)}(\mathbf{W}^{(L)} \mathbf{a}^{(L-1)} + \mathbf{b}^{(L)})
    \end{equation}
    
\end{enumerate}

The combination of these procedures therefore enables DNNs to learn complex patterns and relations in data, hence finding a wide range of applications in many areas of Artificial Intelligence. This feature extraction capability and the ability of pattern recognition within the network are driven both by the depth and breadth of the layers and the nonlinear activation functions in use.

Starting from the early 2010s, DNNs regained momentum with the availability of massive datasets, computing hardware, especially GPUs, and better algorithms. In 2012, a deep neural network-also better known as AlexNet-designed by Alex Krizhevsky, Ilya Sutskever, and Geoffrey Hinton, breasted the winning tape in the ImageNet Large Scale Visual Recognition Challenge by a huge margin, really asserting the result in favor of the efficiency of DNNs on tasks of image classification and recognition~\cite{krizhevsky2012imagenet}. That has been a very important milestone in the timeline of neural network research and also initiated a rapid increase of interest and investments in deep learning technologies across various verticals. The success of AlexNet showed more that deep neural networks, especially the ones using convolutional layers, could have much better scalability to real-world data complexity compared to conventional machine learning models prevalent then. Nevertheless, AlexNet introduced two very significant innovations: ReLU activation functions and dropout. These helped solve some of the main issues in deep network training and, therefore, permitted the creation of even deeper models, overcoming the vanishing gradient problem. That work was followed by the wide adoption of these techniques and further network design improvements, which brought unprecedented advances in medical imaging, self-driving cars, and natural language processing, among many others-basically changing the way these technologies are put into practice.

The success of AlexNet proved not only the power of deep neural networks in handling large and complex data but also how this opened up the avenue leading to the evolution of much more specialized architectures, including Convolutional Neural Networks. As a direct descendant of the earlier DNNs, CNNs leverage layered, hierarchical structures to process spatial and temporal data with unprecedented effectiveness and efficiency. Certain properties of CNNs are designed to take advantage of the spatially local correlation present in images and video data; hence, they are very suited for visual recognition and image processing tasks, as will be presented in this next section.

\subsection*{Convolutional Neural Networks (CNNs)}
\justifying
As the subclass of deep neural networks, CNN usually fits in with the data of grid-like topology, such as an image--one of the simplest 2-D arrays of pixels. Unlike other deep neural networks, CNNs make use of special layers that effectively learn spatial hierarchies and patterns. The usual CNN architecture has the following general mathematical representation:

\begin{figure}[h]
    \centering
    \includegraphics[scale=0.6]{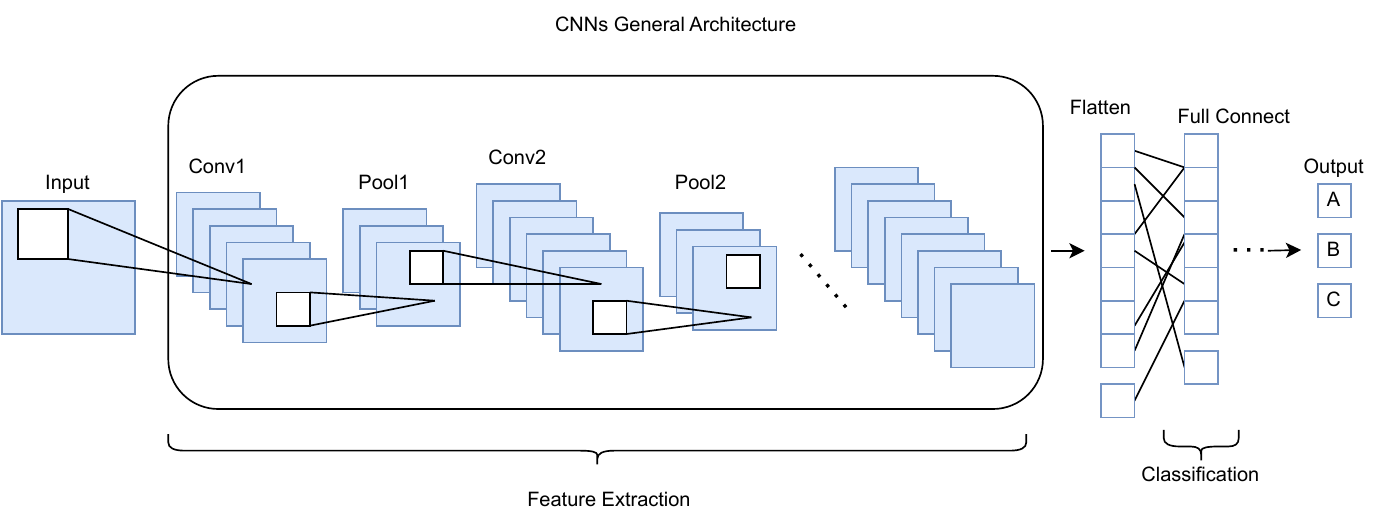}
    \captionsetup{justification=centering}
    \caption[CNN Model Architecture]{\bf General model architecture of a CNN}{This figure illustrates a typical CNN architecture, including convolutional and pooling layers for feature extraction, followed by fully connected layers for classification.}%
    \label{fig:CNN}
\end{figure}

\begin{enumerate}
    \item \textbf{Convolutional Layer:} Applies a convolution operation between the input data $\mathbf{X}$ and a set of learnable filters or kernels $\mathbf{K}$. For an input image $\mathbf{X} \in \mathbb{R}^{H \times W \times C}$ 
    and a filter $\mathbf{K} \in \mathbb{R}^{k \times k \times C}$, the convolution operation for a single filter is defined as:
    
    \begin{equation}
        \mathbf{Z}_{ij} = (\mathbf{K} * \mathbf{X})_{ij} = \sum_{m=1}^{k} \sum_{n=1}^{k} \mathbf{K}_{mn} \mathbf{X}_{i+m-1, j+n-1}
    \end{equation}
    
    where $\mathbf{Z} \in \mathbb{R}^{(H-k+1) \times (W-k+1)}$ is the output feature map, $*$ denotes the convolution operation, $H$ and $W$ are the height and width of the input image, $C$ is the number of channels, and $k$ is the filter size.
    
    \item \textbf{Pooling Layer:} Reduces the spatial dimensions (width and height) of the input volume for the next convolutional layer. A common pooling operation is max pooling with a pooling window of size $p \times p$, which is defined as:
    
    \begin{equation}
        \mathbf{P}_{ij} = \max_{(m,n) \in \mathcal{R}_{ij}} \mathbf{A}_{mn}
    \end{equation}
    
    where $\mathbf{P} \in \mathbb{R}^{(\lfloor H/p \rfloor) \times (\lfloor W/p \rfloor)}$ is the output after pooling, and $\mathcal{R}_{ij}$ is the rectangular region of size $p \times p$ centered at $(i, j)$ in the input $\mathbf{A}$.
    
    \item \textbf{Dropout Layer:} Helps in regularization and preventing overfitting by randomly setting a fraction of input activations to zero during training.
    
    \item \textbf{Fully Connected Layer:} After several convolutional, pooling, and dropout layers, the high-level reasoning in the network is done via fully connected layers. Neurons in a fully connected layer have full connections to all activations in the previous layer, as seen in traditional neural networks. The operation can be mathematically represented as:
    
    \begin{equation}
        \mathbf{y} = \mathbf{W} \cdot \mathbf{a} + \mathbf{b}
    \end{equation}
    
    where $\mathbf{y}$ is the output vector, $\mathbf{W}$ is the weight matrix, $\mathbf{a}$ is the input activations, and $\mathbf{b}$ is the bias vector.

    \item \textbf{Activation Function:} Introduces non-linearity to the network, allowing it to learn complex patterns. A common activation function is the Rectified Linear Unit (ReLU), defined as:
    
    \begin{equation}
        \mathbf{A} = \sigma(\mathbf{Z}) = \max(0, \mathbf{Z})
    \end{equation}
    
    where $\mathbf{A}$ is the output after applying the activation function $\sigma$. The Activation Function typically follows the Convolutional and Fully Connected layers.

\end{enumerate}

This is the architectural sequence in which each layer is built from the processed outputs of earlier layers, and it is what allows CNNs to automatically and adaptively learn spatial hierarchies of features from input images or other grid-like data. This strong structure, illustrated in Fig \ref{fig:CNN}, forms the very basis for its outstanding performance in such tasks as image and video recognition, among other applications entailing detailed spatial analysis.

\subsubsection*{BayesianCNN}
\justifying
The strong backbone of CNNs, especially their prowess at processing and interpreting spatial information, laid a very firm foundation for further innovations into neural network architectures. A recent innovative extension of this family is a Bayesian Convolutional Neural Network, BayesianCNN. While traditional CNNs use point estimates for weights at the time of inference, BayesianCNN introduces probabilistic weights that allow the network to estimate uncertainty in its prediction.

BayesianCNN introduces probabilistic approaches to the weights in the network, enabling the model to express uncertainties. This is approached using {Variational Inference (VI)} and, specifically, through the methodology of {Bayes by Backprop} itself, which approximates intractable true posterior distributions over weights with variational distributions.

VI is an extremely powerful way of approximating Bayesian posterior distributions, which are usually intractable. The key idea of VI is to find an approximate distribution to the true weighted posterior distribution \textit{w} and data, \textit{D} is \( p(w|D) \) with a more tractable distribution \( q_\phi(w) \), where \( q_\phi(w) \) is the variational distribution of the weights, parameterized by \( \phi \). This approximation minimizes the Kullback-Leibler (KL)~\cite{kingma2015variational} divergence between \( q_\phi(w) \) and the true posterior, effectively turning the inference problem into an optimization problem. 


In general, the objective of VI is defined as:
\begin{equation}\label{vi_28}
    \mathcal{L}(\phi) = \mathbb{E}_{q_\phi(w)}[\log p(D|w)] - D_{\mathrm{KL}}(q_\phi(w) \parallel p(w))
\end{equation}
where \( \mathcal{L}(\phi) \) is the variational lower bound, or evidence lower bound (ELBO), the expectation term $\mathbb{E}_{q_{\boldsymbol{\phi}}(w)}\left[\log p(D|w)\right]$ evaluates the likelihood of the data under the current model parameters, promoting the accuracy of predictions. \( p(D|w) \) is the likelihood of the data given the parameters, $p(w)$ is the prior distribution over the weights and \( \mathrm{KL} \) represents the Kullback-Leibler divergence between the variational distribution and the prior.
For the equation \ref{vi_28}, the closer \( \mathrm{KL} \) diverges to `0', minimum \( \mathcal{L}(\phi) \) value is obtained.

So, the variational objective, i.e., the loss function of a Bayesian CNN, is formulated to measure how well the model adheres to the data while also constraining the parameters in a Bayesian framework. This objective combines the likelihood of the observed data given the model's parameters with the KL divergence between the variational distribution of the weights and its prior distribution. This serves as a regularizer that discourages the distance of the variational distribution to the prior knowledge, taken into account to the prior knowledge while keeping overfitting in check by driving the weight distribution to be simple. So, including the data \textit{D},  we can rewrite equation \ref{vi_28} as following:

\begin{equation}\label{eq:bcnn}
\mathcal{L}(\boldsymbol{\phi}; D) = \mathbb{E}_{q_{\boldsymbol{\phi}}(w)}\left[\log p(D|w)\right] - D_{\mathrm{KL}}(q_{\boldsymbol{\phi}}(w|D) \parallel p(w|D))
\end{equation}

In practice, VI is implemented using the Stochastic Gradient Variational Bayes (SGVB) technique, which employs stochastic gradient descent to optimize \( \mathcal{L}(\phi;D) \). The gradients are estimated using samples from \( q_\phi(w|D) \), allowing the use of mini-batch optimization methods common in machine learning. The general expression for a gradient update is given by:
\begin{equation}
    \phi \leftarrow \phi - \eta \nabla_\phi \mathcal{L}(\phi;D),
\end{equation}
where \( \eta \) is the learning rate. The gradients are approximated using:
\begin{equation}
    \nabla_\phi \mathcal{L}(\phi;D) \approx \frac{1}{M} \sum_{i=1}^M \left[ \nabla_\phi \log q_\phi(w^{(i)}|D) (\log p(D|w^{(i)}) - \log q_\phi(w^{(i)}|D) + \log p(w^{(i)}|D)) \right]
\end{equation}
    
with \( w^{(i)} \) sampled from \( q_\phi(w |D)\), and \( M \) being the number of samples used to estimate the gradient.

This method allows for efficient and scalable Bayesian inference, making it applicable to large datasets and complex models typical in modern deep learning applications.

Bayes by Backprop~\cite{shridhar2019comprehensive}, a variational inference method, learns the posterior distribution of the weights \( w \sim q_\phi(w|D) \) in a neural network. It employs backpropagation to sample weights and regularizes them by minimizing a compression cost, known as the variational free energy or the expected lower bound on the marginal likelihood.

Since the true posterior \( p(w|D) \) is typically intractable, an approximate distribution \( q_\phi(w|D) \) is defined to be as close as possible to the true posterior, measured by the KL divergence. The optimal parameters \(\phi_{opt}\) are defined as:

\begin{align}
\phi_{opt} &= \arg \min_\phi \text{KL} [q_\phi(w|D) \parallel p(w|D)] \\
&= \arg \min_\phi \left( \text{KL} [q_\phi(w|D) \parallel p(w)] - 
     \mathbb{E}_{q_\phi(w|\phi)}[\log p(D|w)] + \log p(D) \right),
\end{align}
where the KL divergence is computed as:
\begin{equation}
\text{KL} [q_\phi(w|D) \parallel p(w)] = \int q_\phi(w|D) \log \frac{q_\phi(w|D)}{p(w|D)} \, dw.
\end{equation}

This approach forms an optimization problem with the variational free energy, composed of a complexity cost \( \text{KL} [q_\phi(w|D) \parallel p(w)] \) and a likelihood cost \( \mathbb{E}_{q(w|\phi)}[\log p(D|w)] \). The term \( p(D) \) is defined as the probability of the data \( D \), which represents the marginal likelihood or evidence. The term \( \log p(D) \) can be omitted in optimization as it is constant.

Given the intractability of the KL divergence for exact computation, a stochastic variational method is used. Weights \( w \) are sampled from the variational distribution \( q_\phi(w|D) \), which is more feasible for numerical methods than sampling directly from the true posterior \( p(w|D) \). Thus, we approximate the cost function as:

\begin{equation}
F(D, \phi) \approx \sum_{i=1}^n \left( \log q_\phi(w^{(i)}|D) - \log p(w^{(i)}) - \log p(D|w^{(i)}) \right)
\end{equation}
where \( n \) is the number of draws and \( w^{(i)} \) are samples drawn from \( q_\phi(w|D) \).

This framework has been applied with success in training feedforward and recurrent neural networks, but its application in CNN is unexplored.

The reparameterization trick for convolutional layers in the local manner allows one to perform efficient and computationally accelerated variational inference by shifting the sampling process from weights to layer activations. This method not only accelerates the computation but also brings uncertainty directly into the network activations.

The posterior approximation for the weights is defined as a fully factorized Gaussian distribution, denoted by \( q_\phi(w_{i,j}|D) = \mathcal{N}(\mu_{i,j}, \sigma^2_{i,j}) \) for each weight \( w_{i,j} \) in the set or weight matrix \( W \) where \(i,j \) are two consecutive layers. 
The value refers to the specific weights (\(w_{i,j}\)) sampled from the Gaussian distribution with mean \( \mu_{i,j} \) and variance \( \sigma_{i,j}^2 \). 
The mean and variance are learned during training to express the uncertainty in the model's predictions.
And the nodes are the neurons in the two consecutive layers (\(i,j \)), being connected. This implies that each weight \( w_{i,j} \) is generated by sampling from the Gaussian distribution, expressed as:
\[
w_{i,j} = \mu_{i,j} + \sigma_{i,j} \epsilon_{i,j}, \quad \text{where} \quad \epsilon_{i,j} \sim \mathcal{N}(0, 1).
\]
and $\epsilon_{i,j}$ represents a noise term that follows a standard normal distribution, $\mathcal{N}(0,1)$.

The variational posterior probability distribution $q_\phi(w_{ij}|D)$ is defined as a normal distribution with mean $\mu_{ij}$ and variance $\alpha_{ij}\mu^2_{ij}$, where \(i,j \) are two consecutive layers. The $\alpha_{ij}$ represents a scaling factor that modifies the variance of the normal distribution. Mathematically, it can be expressed as:

\begin{equation}
q_\phi(w_{i,j}|D) = \mathcal{N}(\mu_{i,j}, \alpha_{i,j}\mu^2_{i,j})
\end{equation}

This formulation of the variational posterior probability distribution allows for the implementation of the local reparameterization trick in convolutional layers.

If we consider a standard fully connected neural network containing a hidden layer consisting of 1000 neurons. This layer receives an $M \times 1000$ input feature matrix $A$ from the layer below, which is multiplied by a $1000 \times 1000$ weight matrix $W$, before a nonlinearity is applied, i.e., $B = AW$. We then specify the posterior approximation on the weights to be a fully factorized Gaussian, i.e.,
\begin{equation}\label{eq17}
    q_\phi(w_{i,j}|D) = \mathcal{N}(\mu_{i,j}, \sigma^2_{i,j}) \quad \forall w_{i,j} \in W,
\end{equation}
which means the weights are sampled as
\begin{equation}
    w_{i,j} = \mu_{i,j} + \sigma_{i,j} \varepsilon_{i,j}, \quad \text{where } \varepsilon_{i,j} \sim \mathcal{N}(0, 1).
\end{equation}

Given Equation~\ref{eq17},
it follows that:
\begin{equation}
    q_\phi(b_{m,j} | A) = \mathcal{N}(\gamma_{m,j}, \delta_{m,j}),
\end{equation}
with
\begin{equation}
    \gamma_{m,j} = \sum_{i=1}^{1000} a_{m,i} \mu_{i,j},
\end{equation}
and
\begin{equation}
    \delta_{m,j} = \sum_{i=1}^{1000} a^2_{m,i} \sigma^2_{i,j}.
\end{equation}

The mathematical formulation of this trick in convolutional layers is presented below:

\begin{equation}
b_{j} = A_i \ast \mu_i + \epsilon_j \odot \sqrt{A_i^2 \ast (\alpha_i \odot \mu_i^2)}
\end{equation}

where \(b_{j}\) represents the output activations of the \(j\)-th layer, \(A_i\) is the receptive field matrix of the previous layer \(i\), and \( \mu_i \) denotes the vector of means from the associated weights for layer \(i\). The term \( \alpha_i \) represents the scaling factor for the variance, \( \ast \) denotes the convolution operation, and \( \odot \) represents element-wise multiplication. The noise variable \( \epsilon_j \sim \mathcal{N}(0,1) \) is sampled from a standard normal distribution and helps capture uncertainty. This reparameterization enables the layer to model local uncertainty independently for each activation.




In practical implementation, the network conducts two sequential convolution operations for each layer during the forward pass:

\begin{itemize}
    \item The first convolution, Equation~\ref{eq23}, computes the mean of the output activations using the mean parameters of the weights.
    \item The second convolution, Equation~\ref{eq24}, calculates the variance associated with each activation, thereby integrating uncertainty directly into the network's forward computations.
\end{itemize}

These operations are mathematically expressed as follows:

\begin{align} \label{eq23}
\mu^{(l)} &= w^{(l)} \ast x^{(l-1)} + b^{(l)} 
\end{align}
\begin{align}
\label{eq24}
\sigma^{2(l)} &= \text{Softplus}(w^{(l)}_{\sigma} \ast x^{(l-1)} + b^{(l)}_{\sigma})
\end{align}

Here, \(\mu^{(l)}\) and \(\sigma^{2(l)}\) are vectors where \(w^{(l)}\) and \(b^{(l)}\) are the mean weight and bias parameters for the convolution operation, while \(w^{(l)}_{\sigma}\) and \(b^{(l)}_{\sigma}\) represent the parameters learned for modeling the variance, with Softplus ensuring non-negative variance values.

The Softplus function with a steepness parameter \( \beta \) is a smooth, differentiable transformation used in machine learning to ensure that output values are strictly positive. It is particularly useful for variance parameters in statistical models. The mathematical expression for the Softplus function, incorporating the steepness parameter \( \beta \), is given by:

\begin{equation}
\text{Softplus}_{\beta}(x) = \frac{1}{\beta} \log(1 + e^{\beta x})
\end{equation}

For \( \beta = 1 \), this function reduces to the standard Softplus function:

\begin{equation}
\text{Softplus}(x) = \log(1 + e^x)
\end{equation}

The function gradually approaches zero as \( x \) approaches negative infinity and asymptotically approaches \( x \) as \( x \) goes to positive infinity, resembling the shape of a smoothed hinge. This behavior is useful for avoiding the zero values that can arise from the ReLU function, ensuring numerical stability and non-zero gradients across all input values.

\subsection*{U-Net}
\justifying
While BayesianCNNs were quite robust in handling the uncertainties within deep learning models, there are applications-for instance, medical image segmentation-that require architectures which represent fine-grained spatial hierarchies with high preciseness. U-Net, originally designed for biomedical image segmentation, addresses the requirements by introducing a very unique architecture that remarkably improved the effectiveness of localizing and segmenting objects within an image. The following section describes the U-Net architecture, its design, and operation, including reasoning as to why it is particularly suited to tasks such as segmenting complex anatomical structures in medical imaging.
\vspace{0.1cm}

\textbf{U-Net Architecture}


\begin{figure}[!h]
    \centering
    \includegraphics[scale=0.5]{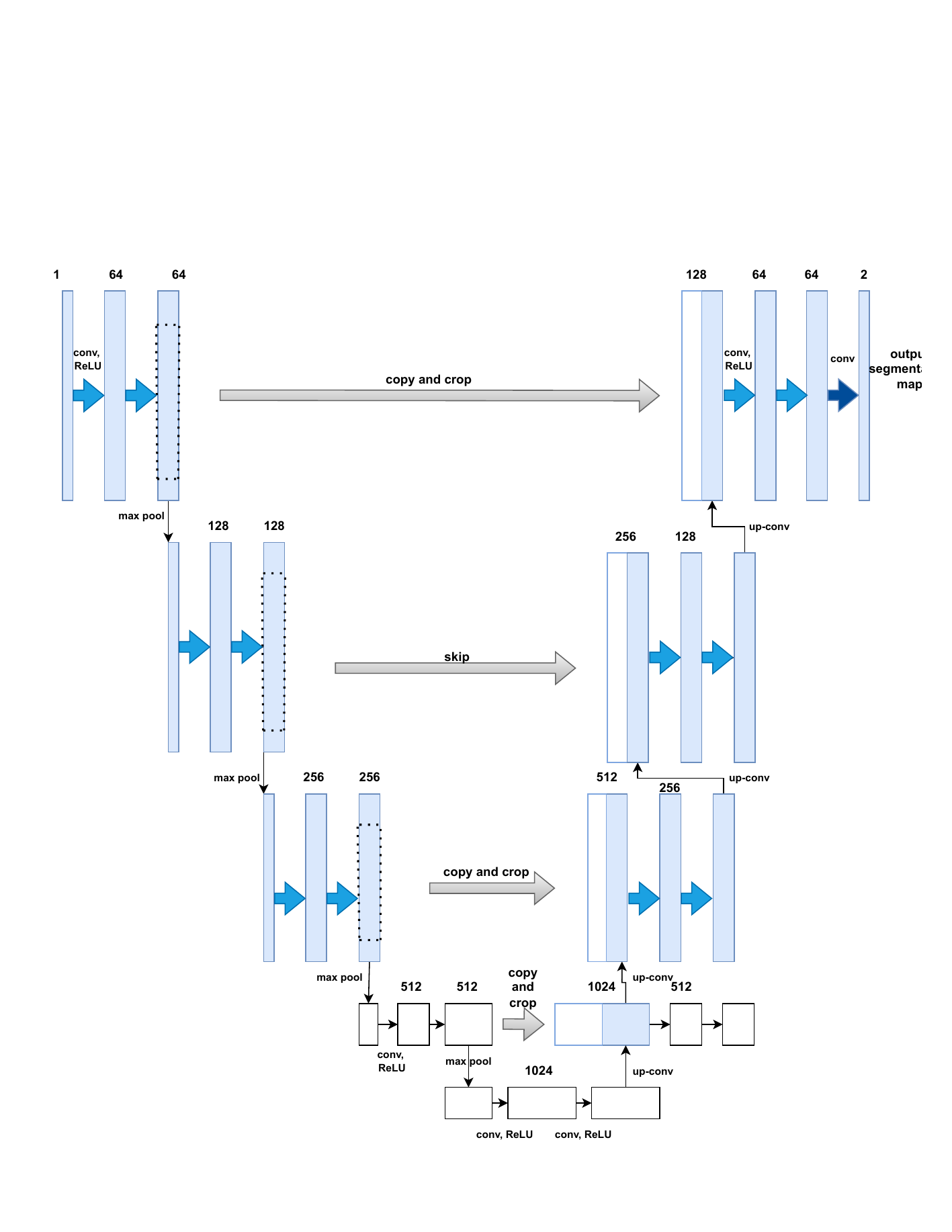}
    \captionsetup{justification=centering}
    \caption[U-Net Model Architecture]{\bf General model architecture of an U-Net}{The figure illustrates the U-Net architecture, consisting of a contracting path (left) for feature extraction through convolution and max-pooling, and an expansive path (right) for precise localization using up-convolutions and skip connections.}%
    \label{fig:unet_0}
\end{figure}

U-Net's architecture at Fig \ref{fig:unet_0} is an advanced encoder-decoder network that uses skip connections to enhance feature integration across the network, facilitating precise localization and context incorporation.
\vspace{0.2cm}

\textbf{Encoder Path}

The encoder, or the contraction path, comprises several convolutional and pooling layers designed to capture the hierarchical features of the input image. Each layer in the encoder can be represented as:
\begin{equation}
    C_i = P(\text{ReLU}(C(\mathbf{W_{i-1}} * X_{i-1} + \mathbf{b_{i-1}})))
\end{equation}
where \(C_i\) is the output of the \(i\)-th convolutional layer, \(P\) denotes a max-pooling operation, \(ReLU\) is the rectified linear activation function, the letter 
\(C\) stands for the convolution operation, which is a fundamental building block of the U-Net model, allowing it to extract hierarchical features from the input images. The \(\mathbf{W_{i-1}}\) and \(\mathbf{b_{i-1}}\) are the weights and biases of the convolutional layer, \(X_{i-1}\) is the input to the \(i\)-th layer, and \(*\) represents the convolution operation.
\vspace{0.2cm}

\textbf{Decoder Path}

The decoder, or the expansion path, upsamples the feature maps to enable precise localization. Each layer in the decoder is defined as:
\begin{equation}
    D_i = \text{ReLU}(C(\mathbf{W'_i} * U(D_{i-1}) + \mathbf{b'_i}) + S_i)
\end{equation}
where \(D_i\) is the output of the \(i\)-th decoder layer, \(U\) denotes an upsampling operation, \(\mathbf{W'_i}\) and \(\mathbf{b'_i}\) are the weights and biases for the decoder, \(S_i\) is the skip-connected feature from the corresponding encoder layer, and \(C\) is the convolution operation.
\vspace{0.2cm}

\textbf{Skip Connections}

Skip connections help recover spatial information lost during downsampling:
\begin{equation}
    S_i = X_{n-i}
\end{equation}
where \(X_{n-i}\) is the output of the \((n-i)\)-th encoder layer directly concatenated to the \(i\)-th layer in the decoder path, facilitating the integration of low-level features with high-level ones.

\section*{Results}
\justifying
In this section we show the experimental setup, the details of the implementation, and the results of the extensive evaluation of our 3 DL models on the OASIS dataset, which has been balanced by using the SMOTE-Tomek ``\nameref{smotetomek}" technique to mitigate issues related to class imbalance. Each model's performance is tested based on a variety of metrics that include accuracy, precision, recall, f1\_score, and area under the ROC curve (AUC), which together present a full view of the capabilities and limitations of dealing with balanced neuroimaging data. The following subsections will describe in detail the balancing of the data using SMOTE-Tomek, the DL models, their training procedures, and the resultant analyses.

\subsection*{Data Pre-processing \& Data Balancing with SMOTE-Tomek} \label{smotetomek}
\justifying
Synthetic Minority Over-sampling Technique (SMOTE) and Tomek links to solve class imbalance problem in machine learning data. SMOTE, initially introduced by Chawla et al.~\cite{chawla2002smote} in 2002, is a well-regarded method for oversampling the minority class by creating synthetic samples rather than simply duplicating existing samples. This method is useful to achieve better generalization than memorization during the training of the model.

(In contrast), Tomek links~\cite{tomek1976two} are employed for undersampling, by detecting pairs of closely related instances of opposite class and removing majority class instances from such pairs. This technique, introduced by Tomek, who first described it in 1976~\cite{tomek1976experiment}, is especially useful for cleaning overlapping between class data points improving the performance of the classifier by making the decision boundary more discriminative.

It increases minority class representation via synthetic sample generation (SMOTE), and in the mean time refines training dataset that removes Tomek links that are either noise or borderline samples. The integration of this combination of techniques leads to a more balanced dataset with which to train the model, which leads to improved model performance, especially in terms of both accuracy and the stability of the classification boundaries. This dual approach not only tackles the problem of imbalanced classes more effectively but also enhances the quality of the synthetic samples produced, leading to improved learning outcomes in predictive modeling tasks.

\subsection*{Results of CNN, BayesianCNN \& U-Net Models}
\justifying
The models under investigation include the CNN based Alzheimer Disease Detection Network (ADD-Net) model, BayesianCNN model \& U-Net model.
\subsubsection*{Architecture \& Result of ADD-Net}
\label{gcam}
\justifying
In this section, we present the methodology of the ADD-Net~\cite{fareed2022add} for early detection of AD can be found in Fig \ref{fig:addnet_1} in their original setting:

\begin{figure}[h]
\centering
  \includegraphics[scale=0.7]{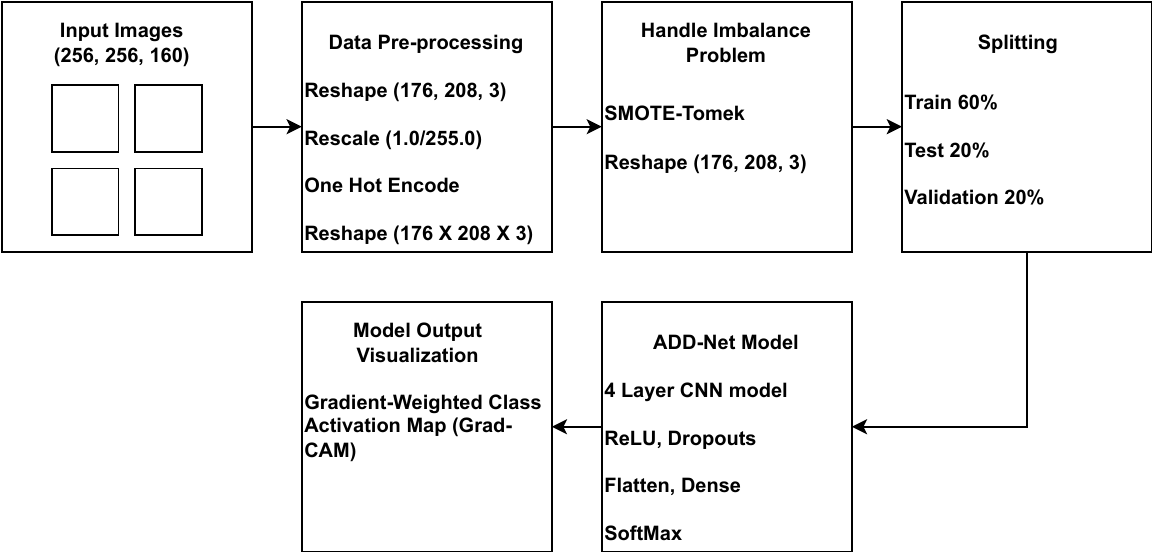}
  \captionsetup{justification=centering}
  \caption[ADD-Net Methodology]{\bf Methodology of the ADD-Net}{The workflow starts with input images, the data undergoes pre-processing, including reshaping, rescaling, and one-hot encoding. Data imbalance is addressed using SMOTE-Tomek, followed by data splitting into training, testing, and validation sets. It consists of a 4-layer CNN with ReLU activations, dropout layers, and softmax for classification. Finally, model output is visualized using Grad-CAM.}%
  \label{fig:addnet_1}
\end{figure}

Muhammad et al.~\cite{fareed2022add} utilized the Kaggle multiclass brain MRI image dataset~\cite{kaggle_alzheimers}, which lacks reliability for scientific publications due to the absence of documented data sources, pre-processing procedures, and labeling methods. Like any other dataset, this data also have major class imbalance problem. The authors employed a synthetic oversampling technique to evenly distribute images among classes and mitigate the class imbalance issue. The ADD-Net model is constructed from scratch and includes four convolutional blocks, each comprising a Rectified Linear Unit (ReLU) activation function, a 2D average pooling layer, two dropout layers, two dense layers, and a SoftMax classification layer. Average pooling calculates the average value of the elements within a specified pooling window. It emphasizes the average presence of features within the feature of interest. 

Gradient-weighted Class Activation Mapping (Grad-CAM) ``\nameref{gradcam}" is used to generate heatmaps on the brain images. Grad-CAM is a technique to visualize regions in an input image that contribute most to the predictions made by a CNN. The later layers in CNNs capture more abstract and high-level features that are more semantically meaningful for decision-making. The last convolutional layers retain spatial information about where in the image certain features appear. These heatmaps visually indicate the most relevant regions for the model classifications, hence providing insights into the model's decision-making process that may help medical professionals to understand the model's predictions.

In this context, we combined SMOTE-Tomek with Grad-CAM on the proposed ADD-Net model for the OASIS dataset. We added batch normalization, which is used to stabilize the learning process and drastically reduce the number of training epochs that deep networks need. The dropout rate was changed to 0.03 after tuning various hyperparameters such as batch size, number of epochs, and learning rates. First, we tuned the model concerning different dropout values preventing overfitting and simultaneously not degrading model performance. This fine-tuning method allowed us to find the optimal dropout rate of 0.03, where regularization and model complexity are well balanced. Further, we set the learning rate to 0.01 to allow smoother convergence. The 80-20 ratio in the division of data for our models was followed: 80\% of the data will be used for training purposes, while 20\% is kept for model testing to ensure that our model performances sound.
We also increased the depth (adding more convolutional layers) which helps the network learn more complex features at various levels of abstraction.
Instead of using ReLU, we consider using advanced activation functions like LeakyReLU which helps in preventing the ``dying ReLU" problem. The ``dying ReLU" problem~\cite{lu2019dying} refers to a situation in a neural network where neurons using the ReLU activation function stop participating in the learning process—effectively ``dying." This occurs because the ReLU function outputs zero for any negative input and only passes values through unchanged when they are positive. If a neuron's weights change during training in such a manner that the weighted sum of its inputs is always negative, the output of the ReLU will be zero every time. This, in turn, implies that the gradient through that neuron during backpropagation will also be zero, which means the weights of that neuron stop updating altogether. Also, different dropout rates are used to find the optimal setting for preventing overfitting the new data with the model. After incorporating these modifications with the OASIS, we get 93.96\% accuracy along with 93.66\% F-1 score, Recall 93\%  and Precision 93\%. 

\subsubsection*{Architecture \& Result of BayesianCNN} 
\begin{figure}[h]
  \centering
  \includegraphics[scale=0.6]{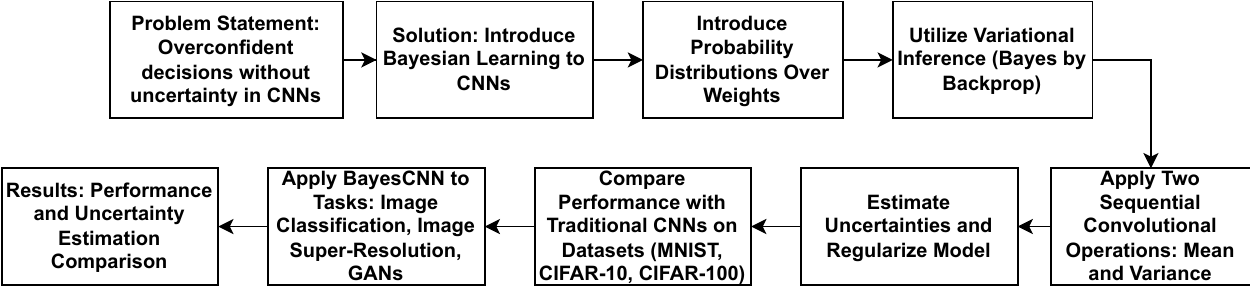}
  \captionsetup{justification=centering}
  \caption[BayesianCNN Methodology]{\bf Methodology of the BayesianCNN}{This diagram addresses overconfidence in traditional CNNs by introducing Bayesian learning. The method applies probabilistic distributions over weights and utilizes variational inference to estimate uncertainties. The model undergoes sequential convolutional operations for mean and variance, followed by comparison with traditional CNNs on various datasets.}%
  \label{fig:bcnn_1}
\end{figure}

\justifying
The authors Kumar et al. introduce BayesianCNN ~\cite{shridhar2019comprehensive} along with Bayes by Backprop can be visualized here in Fig \ref{fig:bcnn_1}.

BayesianCNN introduces probability distributions over the weights of a CNN. It is implemented using Variational Inference with the ``Bayes by Backprop" technique for approximating the true posterior distribution with a variational distribution. The approach mainly revolves around the local reparameterization trick for convolutional layers, translating global uncertainty into local and independent uncertainty across samples and improving both computational efficiency and robustness of gradient estimates. The architecture applies two convolutional operations consecutively for each layer: one to calculate the mean and one for the variance of that particular layer's output. The two-operation framework enables the network to capture uncertainty effectively in the predictions. For this Bayesian learning, as every weight parameter now has mean and variance, model pruning strategies may be introduced to handle the increase in model parameters. It does this by reducing the number of filters by half and utilizing L1 norm for sparsity. This keeps the overall model size manageable without having any performance compromise.
Datasets used here are MNIST~\cite{lecun_mnist_1998}, CIFAR-10~\cite{krizhevsky_cifar10_2009} and CIFAR-100~\cite{krizhevsky_cifar100_2009}.

The Bayesian CNN architecture modifies this by introducing two kinds of convolutional optimizations, one for the estimation of the mean and another one for the variance of weights. This allows the dual approach to be encoded within the network itself and be further processed within the network layers.

The network configuration for the BayesianCNN is shown below:
\begin{itemize}
    \item Two convolutional layers for mean and variance estimation, followed by a max-pooling layer.
    \item Fully connected layers at the end of the network structure.
    \item Activation functions: ReLU for the convolutional layers and Softplus for the variance estimation to ensure non-negative variance.
\end{itemize} 

Its model training utilizes the Adam optimizer that performs better in sparse gradient-related computations and in adaptively handling the learning rate of the model. For example, the training uses an Adam optimizer known for effectiveness when handling sparse gradients as well as making adaptive learning rate adjustments. Finally, this model is tuned many times with a set of hyperparameters; hence, its learning rate has to be adjusted to 0.001. This value is chosen such that it can present a smooth convergence during the training without any threat of oscillation or overshooting the minimum. After testing different learning rates and observing the performances of the model, a value of 0.001 was determined as best to realize accuracy with ensured stability in the trainees.
This also smooths convergence without overshooting the minima. Regarding the loss function, the approach is two-fold: the negative log-likelihood loss is applied for managing the classification tasks effectively, while the Kullback-Leibler divergence approximates the weight posterior distribution, integrating a Bayesian perspective into the training process. Train on up to a maximum of 100 epochs so that results are not overfitted. Our model is trained on an 80-20 ratio, meaning that 80\% of the data is used for training, while the rest of it remains for validation to make sure that the model performs well.

So far, similar to the original author's work, we construct the model to evaluate on the OASIS. As the OASIS data also have a class imbalance problem, we handle it by conducting the SMOTE-Tomek data balancing technique which was discussed in section~``\nameref{smotetomek}". After overcoming this problem, we modify the code for our new data and adjust the hyper-parameters in order to run the code. So in our model, the number of epochs are set to 100, Early Stopping with patience 10 has been used, the initial learning rate is set to 0.001, the number of workers for data loading is 4, and the size of the validation set has been set at 20\% of the total dataset with a batch size for training set to 256 samples per batch accordingly. Below, early stopping is implemented using an EarlyStopping callback from TensorFlow's Keras library. Within this, the callback is set to monitor val\_loss during training. In case there is no improvement in validation loss within a given consecutive number of epochs-which in this example has been set to a patience of 10 epochs-an early stop to the training will be issued. Another important parameter is that restore\_best\_weights is set to True; this means immediately after the early stopping happens, the model weights get restored to the state taken from the epoch with minimum validation loss during training. This step can help to avoid overfitting by stopping the training when the model starts to overfit the training data, and it also will make sure that the last model is the one that performed best on the validation set. After these changes with the kaggle dataset, we get the accuracy of 97.49\% along with F-1 score 96.09\%, Recall 97\% and Precision 97\%. 

\subsubsection*{Architecture \& Result of U-Net} 
\begin{figure}[h]
  \centering
  \includegraphics[scale=0.7]{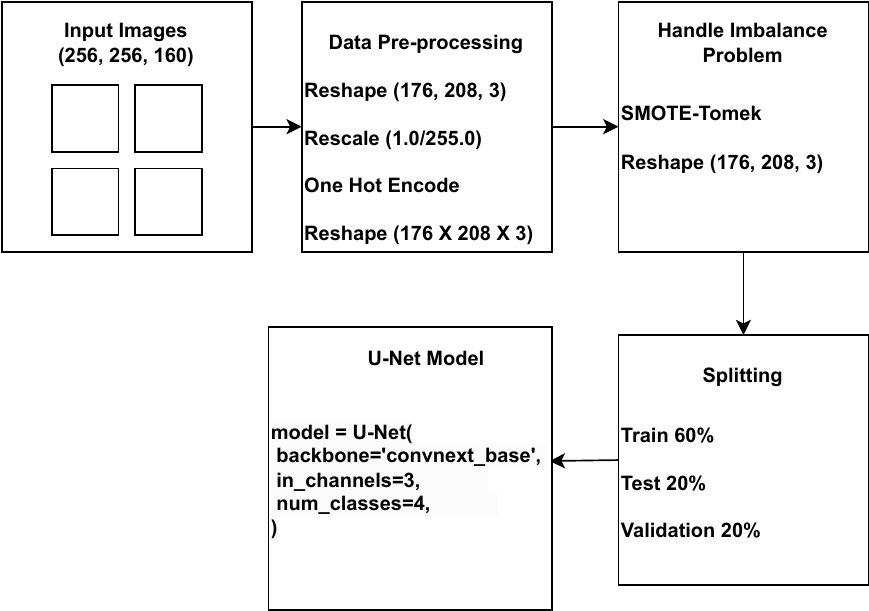}
  \captionsetup{justification=centering}
  \caption[U-Net Methodology]{\bf Methodology of the U-Net}{This figure outlines the U-Net model's workflow for processing MRI images from the OASIS dataset. It includes data pre-processing steps such as reshaping, rescaling, and one-hot encoding. The data imbalance is handled using the SMOTE-Tomek technique. The dataset is then split into training (60\%), testing (20\%), and validation (20\%) sets. This model uses a convolutional base with 3 input channels and 4 output classes for segmentation tasks.}%
  \label{fig:unet_1}
\end{figure}

\justifying
The basic structure of U-Net can be found here \ref{fig:unet_0}. The authors Zhonghao Fan et al. introduce U-Net~\cite{fan2021u} for MRI images. We construct our U-Net model which can be visualized at Fig \ref{fig:unet_1}.

3D T1-weighted MRI images were used from the ADNI database. The authors first pre-processed the data for removing the skull using FreeSurfer, then down-sampling the images to reduce their resolution, clipping to remove non-brain parts, and finally normalizing the intensity of the images so that all have the same brightness and contrast for different scans.

A U-net-based architecture is employed which is effective for analyzing medical images \cite{ronneberger2015u}. It has successive down-sampling and up-sampling layers connected by skip connections in between in order to preserve the spatial hierarchies. The authors adapted the U-net for the classification task from segmentation tasks by adding fully connected layers and softmax activation at the end.

Backpropagation was conducted with the optimizer set as Adam. Batch normalization and dropout were utilized to enhance convergence and prevent overfitting, respectively. The training was carried out with 80\% of the data, and its validation was done with the remaining 20\%. Furthermore, five-fold cross-validation was utilized to ensure that the models are robust. It essentially divides the dataset into five subsets, trains the model on four and validates it on the remaining one, and repeats this five times for all subsets. This kind of elaborate evaluation method reduces the impact of data variance and provides a much closer estimation toward the generalization capability of the model.

The 3D Gradient-weighted Class Activation Mapping (Grad-CAM) technique was adopted for visualizing which parts of the brain MRI scans were most indicative of Alzheimer's disease, hence giving an insight into what the model is focusing on when it made the predictions.

We construct our U-net model, which is configured with a number of important parameters in enhancing its performance in an image segmentation task. The base for the model is ConvNeXt since it can efficiently extract features from medical image data. "convnext\_base" is one of the basic configurations and usually has a good trade-off between complexity and performance. It is designed to handle everything from the simplest classification tasks to complex scenarios like medical image segmentation or object detection, hence finding a very versatile implementation in many various backbone vision machine learning models, such as the U-Net. Inputs are 3-channel images and thus can handle color channels in RGB. It categorizes the segmented regions into four distinct classes based on the complexity and the level of granularity of the segmentation tasks at hand.

For training, U-net uses the Adam optimizer, as this is efficient for sparse gradients and adaptive in noisy training. Cross-entropy loss evaluates the performance of classification output, whose probabilities, in the best case, range from 0 to 1. The initial learning rate is set to 0.001, and a learning rate scheduler is used to dynamically adjust the learning rate in pursuit of better validation loss results. This is fitted with a batch size of 8, which is a strategic choice between computational memory demands and stability of training updates. While it trains up to a maximum of 100 epochs, an early stop mechanism that stops the training process when there is no further improvement in validation accuracy acts to prevent overfitting.

In the U-Net architecture, the reduction rate is 0.9, while the dropout for regularization of the model is done with a high probability of 0.9 while training. Also, adding a small value epsilon to prevent division by zero.

Also the authors did not consider balancing the data nor used OASIS data. We solve the data imbalance problem through SMOTE-Tomek ``\nameref{smotetomek}" and use the OASIS data to evaluate the model which we constructed and modified for our OASIS dataset.

Accuracy, sensitivity, specificity, F1-score, and AUC were used to quantify the performance of the network. Extensive ablation studies were conducted to appraise the contribution of different components, including skip connections and deep supervision, toward diagnostic accuracy. After integrating these modifications with OASIS, an accuracy of 84.91\%, F-1 score of 84\%, Recall of 82\%, and Precision of 87\% were obtained.

\subsection*{Overall Comparison among the DL models}
\justifying

This section presents the performance of a range of deep learning models when applied to the diagnosis of AD through MRI scans.

Some of these models' performances and configurations on the OASIS dataset are contrasted with each other as shown in Table~\ref{tab:model_performance1}. The ADD-Net model, trained using the SMOTE-Tomek hybrid resampling technique, realized a test accuracy of 94.16\% with hyperparameters including a dropout rate of 0.3, batch size of 16, and learning rate of 0.01. Without the application of the SMOTE-Tomek hybrid resampling, the ADD-Net model reached a test accuracy of 92.35\%. While a dropout of 0.3 and batch size of 16 but at a lower learning rate of 0.001, the test accuracy level for U-Net and SMOTE-Tomek U-Net is 84.91\% and 83.18\%, respectively.

\begin{table}[!ht]
\begin{adjustwidth}{-1.5in}{0in}
\caption[Performance and Configuration]{Performance and Configuration Comparison on OASIS}
\label{tab:model_performance1}
\small 
\begin{tabular}{lcc}
\toprule
\textbf{Model Name} & \textbf{Hyperparameters} & \textbf{Test Accuracy}  \\
\midrule
ADD-Net (with SMOTE-Tomek)  & Dropout=0.3, Batch size=16, LR=0.01 & 94.16\%  \\ 
ADD-Net (without SMOTE-Tomek) & Dropout=0.3, Batch size=16, LR=0.01 & 92.35\% \\ 
\midrule 
\textbf{BayesianCNN (with SMOTE-Tomek)} & \textbf{Dropout=0.1, Batch size=16, LR=0.001} & \textbf{95.03\%} \\
BayesianCNN (without SMOTE-Tomek)  & Dropout=0.1, Batch size=16, LR=0.001 & 91.20\% \\ 
\midrule
U-Net (with SMOTE-Tomek)  & Dropout=0.3, Batch size=16, LR=0.001 & 84.91\% \\ 
U-Net (without SMOTE-Tomek) & Dropout=0.3, Batch size=16, LR=0.001 & 83.18\% \\
\bottomrule
\end{tabular}
\end{adjustwidth}
\end{table}


Overall, table \ref{tab:model_performance1} provides insights into the performance and configurations of different models on different datasets.

Table \ref{tab:model_performance3} summarizes the outcomes in terms of Test Accuracy, F1 Score, Recall, and Precision metrics for three models, ADD-Net, BayesianCNN and U-Net. Each model was rigorously evaluated to ascertain its efficacy in accurately identifying diagnostic features from the imaging data. The results highlight the strengths and limitations of each model, providing insights into their practical applicability in clinical settings.

\begin{table}[!ht]
\begin{adjustwidth}{-0.9in}{0in}
\caption[Performance Evaluation]{Performance Evaluation of DL Models on OASIS}
\label{tab:model_performance3}
\small 
\begin{tabular}{lcccc}
\toprule
\textbf{Model Name} & \textbf{Test Accuracy} & \textbf{Recall} & \textbf{Precision} & \textbf{F1-Score} \\
\midrule
ADD-Net (with SMOTE-Tomek) & 94.16\% & 0.92 & 0.95 & 0.93 \\
ADD-Net (without SMOTE-Tomek) & 92.35\% & 0.90 & 0.93 & 0.91 \\ 
\midrule
\textbf{BayesianCNN (with SMOTE-Tomek)} & \textbf{95.03\%} & \textbf{0.93} & \textbf{0.95} & \textbf{0.94} \\
BayesianCNN (without SMOTE-Tomek) & 91.20\% & 0.90 &  0.91 & 0.89\\ 
\midrule
U-Net (with SMOTE-Tomek) & 84.91\% & 0.82 & 0.87 & 0.84 \\
U-Net (without SMOTE-Tomek) & 83.18\% & 0.80 & 0.85 & 0.82 \\ 
\bottomrule
\end{tabular}
\end{adjustwidth}
\end{table}

\subsection*{Masking}
\justifying
Masking, when applied to U-net, means applying a mask which allows only some of the pixels in the output image to contribute towards the loss during training. This is important in medical images where only some regions are relevant, say, tissues or organs, and other areas should be ignored. This more acts like the ground truth. Masking will ensure that the model pays attention to only relevant parts of an image in the process of yielding more accurate, highly relevant segmentation results.

One limitation concerning our U-Net model is that we have skipped the masking part here. The reason behind this is that most of the time, masking is done with manual FLAIR abnormality through expert medical officials. If proper masking is not done, then the model may learn the irrelevant features of the image that decrease the precision in segmentation. It overfits to noise and irrelevant details, thinking that these are essential features, hence compromising the generalization capability on unseen data.

Also, if the same images are used as inputs and masks without actual masking, then it will lead the model to learn everything about the image, which may be irrelevant. It would mess up the model by telling it to focus on which features and ultimately deteriorate the performance it gives for segmentation as it couldn't highlight the important boundaries and features during the training.

\section*{Discussion and Conclusion}
\justifying
The study offers a comparative study of two different models, ADD-Net and BayesianCNN, by using SMOTE-Tomek balanced datasets. An interpretable network in which Grad-CAM helps identify the key areas responsible in brain scan images characterizes ADD-Net. Batch normalization, adding further convolution layers, and the dropout enhancement have yielded greater accuracy and robustness with some modifications. BayesianCNN addresses uncertainty related to medical images through variational inference and dual convolution operations on the other side. While promising, its application currently is confined to the Kaggle brain MRI dataset; it needs further adaptation to the OASIS dataset for wider clinical applications.

Both models have challenges that include computational demands and a risk of overfitting due to synthetic oversampling techniques such as SMOTE-Tomek. How well these issues are managed determines the accuracy and reliability of the models in real-world applications.

BayesianCNN in conjunction with SMOTE-Tomek yielded better performances compared to both ADD-Net and U-Net since it handled imbalanced data effectively enough to yield improved sensitivity and specificity and, therefore, stands out as a potential strong candidate for early detection. This will underline the transformative potential of deep learning models combined with advanced techniques in medical diagnostics. In the near future, attempts at improvement on these models, different methods of masking within 3D U-Net and Attention U-Net will be adopted; applying Grad-CAM on each to interpret its results with further performance optimization for the task of detecting AD. Demographic information will likely have an even better representation for this aspect since no variable was associated with demographic characteristics in relation to the current work. This would further enhance the model's applicability across diverse patient populations. We also deployed a web application publicly for research purposes that will allow broader access, thus supporting early diagnosis of Alzheimer's on a larger scale.

\section*{Supporting information}
\label{gradcam}

As we already discussed in Section \nameref{gcam}, the Gradient-weighted Class Activation Mapping (Grad-CAM) is used to generate heatmaps on the brain images. These heatmaps help in visualizing which parts of the brain images the model is focusing on for its predictions, potentially revealing regions of interest for further analysis or interpretation in medical diagnosis. Here, in summary, the color coding across all Grad-CAM images is consistent:
\begin{itemize}
    \item Red: Most important regions.
    \item Yellow: Moderately important regions.
    \item Green and Blue: Less important regions.
\end{itemize}

The following visualizations illustrate the brain areas the models focus on for their predictions, highlighting potential regions of interest for further analysis or interpretation in medical diagnosis.

\begin{figure}[H]
\centering
\includegraphics[scale=0.7]{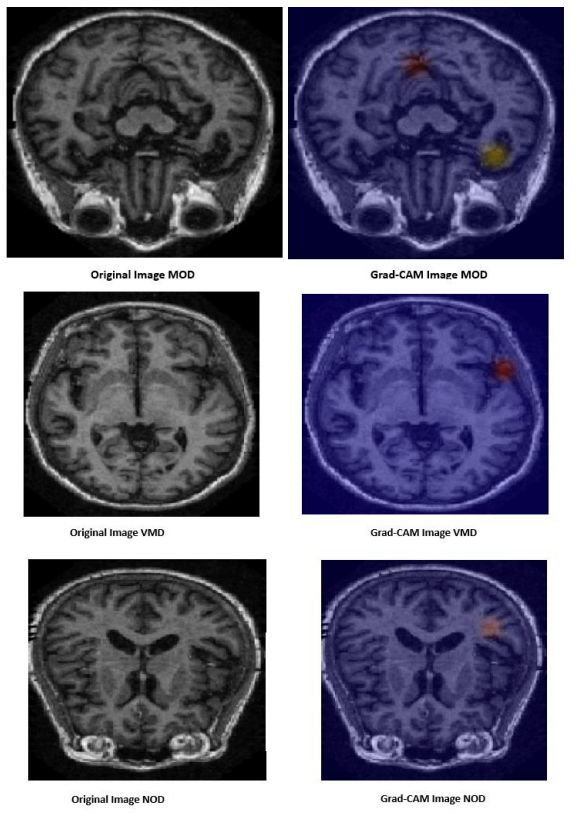}
\captionsetup{justification=centering}
\caption[ADD-Net Grad-CAM]{\bf CNN based ADD-Net on OASIS data Grad-CAM images}
\label{S1_Fig}
\end{figure}

\paragraph*
\justifying
These heatmaps provide a clear view of the region in the brain images that the model is paying attention to for making the predictions, probably the region of interest for further analysis or interpretation in medical diagnosis.

\begin{figure}[H]
\centering
\includegraphics[scale=0.7]{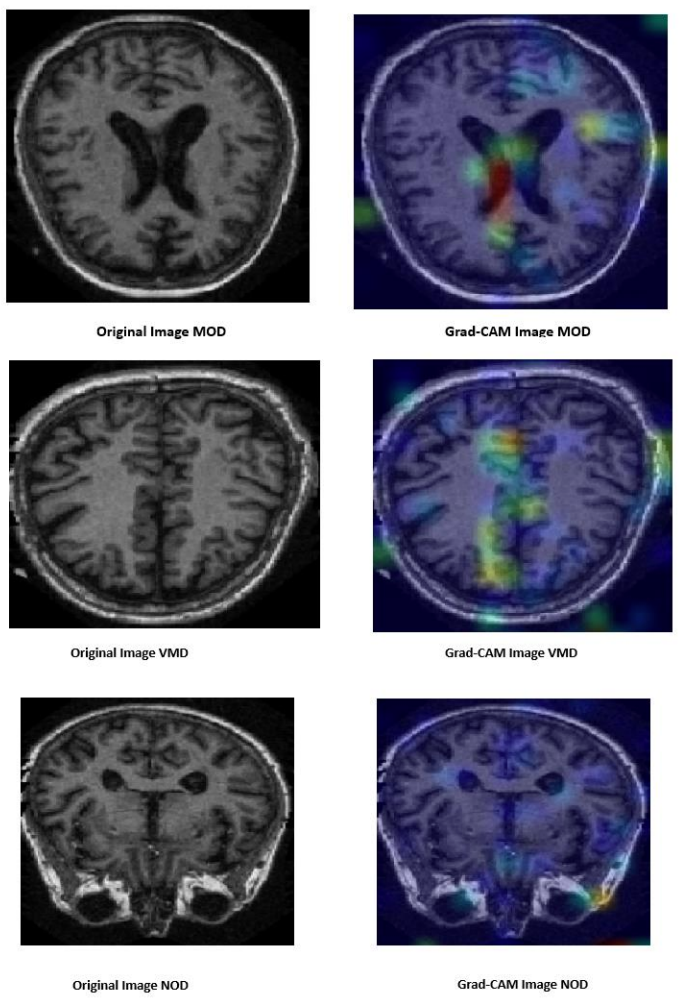}
\captionsetup{justification=centering}
\caption[BayesianCNN Grad-CAM]{\bf BayesianCNN model on OASIS data Grad-CAM images}
\label{S2_Fig}
\end{figure}

\paragraph*
\justifying
These Grad-CAM visualizations from the BayesianCNN model highlight the areas of focus in the brain images. These regions are of particular interest for further medical analysis and diagnosis.

\begin{figure}[H]
\centering
\includegraphics[scale=0.7]{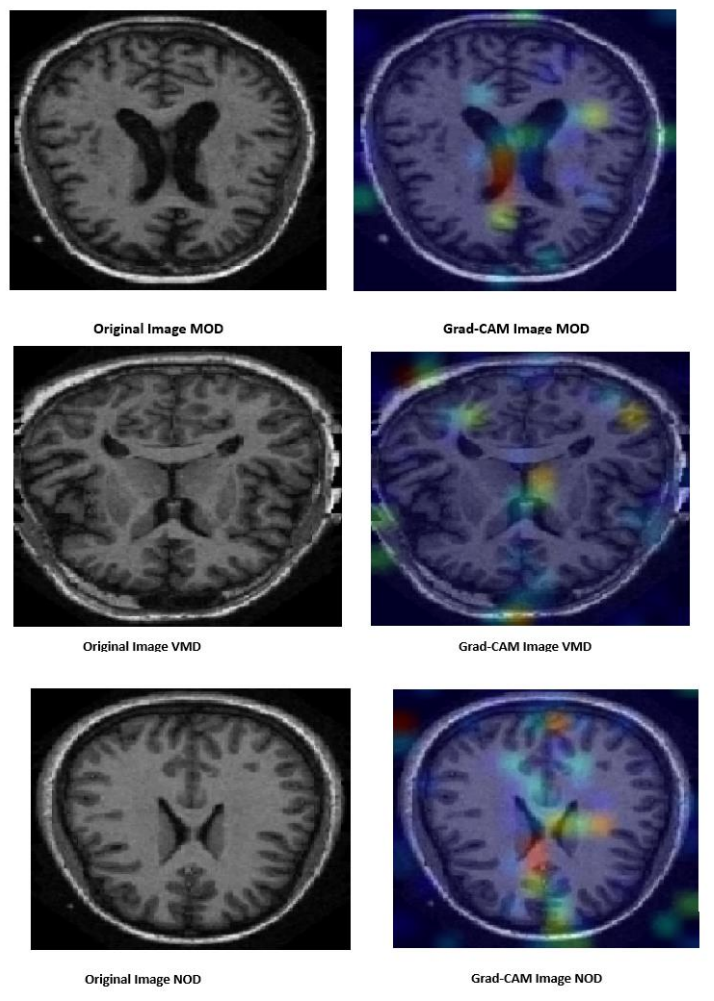}
\captionsetup{justification=centering}
\caption[U-Net Grad-CAM]{\bf U-Net model on OASIS data Grad-CAM images}
\label{S3_Fig}
\end{figure}

\paragraph*
\justifying
The U-Net model’s Grad-CAM heatmaps demonstrate where the model focuses its attention for predictions, aiding in understanding the model's decision-making process.

\clearpage 

\nolinenumbers


\begin{thebibliography}{99} 

\bibitem{WHO2020}
World Health Organization (WHO).
\newblock {{Dementia}}.
\newblock 2020.
\newblock Available from: \url{https://www.who.int/news-room/fact-sheets/detail/dementia}. Accessed 2024.

\bibitem{Garza:2011fk}
Garza G.
\newblock {{Mathematics of Synthetic Aperture Radar Imaging}}.
\newblock [Master's thesis]. The University of Texas-Pan American; 2011.

\bibitem{fareed2022add}
Fareed MMS, Zikria S, Ahmed G, Mahmood S, Aslam M, Jillani SF, et al.
\newblock {{ADD-Net: An effective deep learning model for early detection of Alzheimer's disease in MRI scans}}.
\newblock IEEE Access. 2022;10:96930--96951.

\bibitem{shridhar2019comprehensive}
Shridhar K, Laumann F, Liwicki M.
\newblock {{A comprehensive guide to Bayesian convolutional neural networks with variational inference}}.
\newblock arXiv preprint arXiv:1901.02731; 2019.

\bibitem{blundell2015weight}
Blundell C, Cornebise J, Kavukcuoglu K, Wierstra D.
\newblock {{Weight uncertainty in neural networks}}.
\newblock In: International conference on machine learning. PMLR; 2015. p. 1613--1622.

\bibitem{graves2011practical}
Graves A.
\newblock {{Practical variational inference for neural networks}}.
\newblock Advances in neural information processing systems. 2011;24.

\bibitem{fan2021u}
Fan Z, Li J, Zhang L, Zhu G, Li P, Lu X, et al.
\newblock {{U-net based analysis of MRI for Alzheimer's disease diagnosis}}.
\newblock Neural Computing and Applications. 2021;33:13587--13599.

\bibitem{ronneberger2015u}
Ronneberger O, Fischer P, Brox T.
\newblock {{U-net: Convolutional networks for biomedical image segmentation}}.
\newblock In: Medical image computing and computer-assisted intervention--MICCAI 2015: 18th international conference, Munich, Germany, October 5-9, 2015, proceedings, part III 18. Springer; 2015. p. 234--241.

\bibitem{lecun2015deep}
LeCun Y, Bengio Y, Hinton G.
\newblock {{Deep learning}}.
\newblock Nature. 2015;521(7553):436--444.

\bibitem{marcus2010open}
Marcus DS, Fotenos AF, Csernansky JG, Morris JC, Buckner RL.
\newblock {{Open access series of imaging studies: cross-sectional MRI data in nondemented and demented older adults}}.
\newblock Journal of cognitive neuroscience. 2010;22(12):2677--2684.

\bibitem{oasis_brains}
OASIS: Open Access Series of Imaging Studies.
\newblock {{OASIS Brains}}. Available from: \url{https://www.oasis-brains.org/#data}. Accessed 2024.

\bibitem{morris1993clinical}
Morris JC.
\newblock {{The Clinical Dementia Rating (CDR): current version and scoring rules}}.
\newblock Neurology. 1993;43(11):2412--2412.

\bibitem{weiner2013alzheimer}
Weiner MW, Veitch DP, Aisen PS, Beckett LA, Cairns NJ, Green RC, et al.
\newblock {{The Alzheimer's Disease Neuroimaging Initiative: a review of papers published since its inception}}.
\newblock Alzheimer's \& Dementia. 2013;9(5):e111--e194.

\bibitem{alzubaidi2020towards}
Alzubaidi L, Fadhel MA, Al-Shamma O, Zhang J, Santamaría J, Duan Y, et al.
\newblock {{Towards a better understanding of transfer learning for medical imaging: a case study}}.
\newblock Applied Sciences. 2020;10(13):4523.

\bibitem{shereen2021proposed}
Shereen A, Ghali NI.
\newblock {{A proposed recognition system for Alzheimer’s disease based on deep learning and optimization algorithms}}.
\newblock Journal of Southwest Jiaotong University. 2021;56(5).

\bibitem{mohammed2021multi}
Mohammed BA, Senan EM, Rassem TH, Makbol NM, Alanazi AA, Al-Mekhlafi ZG, et al.
\newblock {{Multi-method analysis of medical records and MRI images for early diagnosis of dementia and Alzheimer’s disease based on deep learning and hybrid methods}}.
\newblock Electronics. 2021;10(22):2860.

\bibitem{battineni2021deep}
Battineni G, Chintalapudi N, Amenta F, Traini E.
\newblock {{Deep Learning Type Convolution Neural Network Architecture for Multiclass Classification of Alzheimer's Disease}}.
\newblock In: Bioimaging; 2021. p. 209--215.


\bibitem{eskildsen2013prediction}
Eskildsen SF, Coupé P, García-Lorenzo D, Fonov V, Pruessner JC, Collins DL, et al.
\newblock {{Prediction of Alzheimer's disease in subjects with mild cognitive impairment from the ADNI cohort using patterns of cortical thinning}}.
\newblock Neuroimage. 2013;65:511--521.

\bibitem{duc20203d}
Duc NT, Ryu S, Qureshi MNI, Choi M, Lee KH, Lee B.
\newblock {{3D-deep learning based automatic diagnosis of Alzheimer’s disease with joint MMSE prediction using resting-state fMRI}}.
\newblock Neuroinformatics. 2020;18:71--86.

\bibitem{islam2017novel}
Islam J, Zhang Y.
\newblock {{A novel deep learning based multi-class classification method for Alzheimer’s disease detection using brain MRI data}}.
\newblock In: Brain Informatics: International Conference, BI 2017, Beijing, China, November 16-18, 2017, Proceedings. Springer; 2017. p. 213--222.

\bibitem{szegedy2017inception}
Szegedy C, Ioffe S, Vanhoucke V, Alemi A.
\newblock {{Inception-v4, inception-resnet and the impact of residual connections on learning}}.
\newblock In: Proceedings of the AAAI conference on artificial intelligence. 2017;31(1).

\bibitem{ghazal2022alzheimer}
Ghazal TM, Issa G.
\newblock {{Alzheimer disease detection empowered with transfer learning}}.
\newblock Computers, Materials \& Continua. 2022;70(3):5005--5019.

\bibitem{alom2018history}
Alom MZ, Taha TM, Yakopcic C, Westberg S, Sidike P, Nasrin MS, et al.
\newblock {{The history began from AlexNet: A comprehensive survey on deep learning approaches}}.
\newblock arXiv preprint arXiv:1803.01164; 2018.

\bibitem{sarraf2016classification}
Sarraf S, Tofighi G.
\newblock {{Classification of Alzheimer's disease using fMRI data and deep learning convolutional neural networks}}.
\newblock arXiv preprint arXiv:1603.08631; 2016.

\bibitem{albawi2017understanding}
Albawi S, Mohammed TA, Al-Zawi S.
\newblock {{Understanding of a convolutional neural network}}.
\newblock In: 2017 international conference on engineering and technology (ICET). IEEE; 2017. p. 1--6.

\bibitem{hazarika2021improved}
Hazarika RA, Abraham A, Kandar DK, Maji AK.
\newblock {{An improved LeNet-deep neural network model for Alzheimer’s disease classification using brain magnetic resonance images}}.
\newblock IEEE Access. 2021;9:161194--161207.

\bibitem{krizhevsky2012imagenet}
Krizhevsky A, Sutskever I, Hinton GE.
\newblock {{ImageNet classification with deep convolutional neural networks}}.
\newblock Advances in neural information processing systems. 2012;25.

\bibitem{buntine1991bayesian}
Buntine WL.
\newblock {{Bayesian backpropagation}}.
\newblock Complex systems. 1991;5:603--643.

\bibitem{radiopaedia_alz}
Radiopaedia.
\newblock {{Alzheimer Disease}}.
\newblock Available from: https://radiopaedia.org/articles/alzheimer-disease-1 [Accessed 2024-11-19].

\bibitem{radiology_assistant}
The Radiology Assistant.
\newblock {{Role of MRI in Dementia Imaging}}.
\newblock Available from: https://radiologyassistant.nl/neuroradiology/dementia/role-of-mri [Accessed 2024-11-19].

\bibitem{alz_research_uk}
Alzheimer's Research UK.
\newblock {{All You Need to Know About Brain Scans and Dementia}}.
\newblock Available from: https://www.alzheimersresearchuk.org/news/all-you-need-to-know-about-brain-scans-and-dementia/ [Accessed 2024-11-19].
%

\bibitem{lecun_mnist_1998}
LeCun Y, Cortes C, Burges CJC.
\newblock {{THE MNIST DATABASE of handwritten digits}}. 1998.
\newblock Available from: \url{http://yann.lecun.com/exdb/mnist/}.

\bibitem{krizhevsky_cifar10_2009}
Krizhevsky A.
\newblock {{Learning Multiple Layers of Features from Tiny Images}}. 2009.
\newblock Available from: \url{https://www.cs.toronto.edu/~kriz/learning-features-2009-TR.pdf}.

\bibitem{krizhevsky_cifar100_2009}
Krizhevsky A.
\newblock {{Learning Multiple Layers of Features from Tiny Images}}.
\newblock 2009.
\newblock Available from: \url{https://www.cs.toronto.edu/~kriz/learning-features-2009-TR.pdf}.


\bibitem{kingma2015variational}
Kingma DP, Salimans T, Welling M.
\newblock {{Variational dropout and the local reparameterization trick}}.
\newblock Advances in neural information processing systems. 2015;28.

\bibitem{chen2019s3d}
Chen W, Liu B, Peng S, Sun J, Qiao X.
\newblock {{S3D-UNet: separable 3D U-Net for brain tumor segmentation}}.
\newblock In: Brainlesion: Glioma, Multiple Sclerosis, Stroke and Traumatic Brain Injuries: 4th International Workshop, BrainLes 2018, Held in Conjunction with MICCAI 2018, Granada, Spain, September 16, 2018, Revised Selected Papers, Part II 4. Springer; 2019. p. 358--368.

\bibitem{islam2020brain}
Islam M, Vibashan V, Jose VJM, Wijethilake N, Uppal U, Ren H.
\newblock {{Brain tumor segmentation and survival prediction using 3D attention UNet}}.
\newblock In: Brainlesion: Glioma, Multiple Sclerosis, Stroke and Traumatic Brain Injuries: 5th International Workshop, BrainLes 2019, Held in Conjunction with MICCAI 2019, Shenzhen, China, October 17, 2019, Revised Selected Papers, Part I 5. Springer; 2020. p. 262--272.

\bibitem{chawla2002smote}
Chawla NV, Bowyer KW, Hall LO, Kegelmeyer WP.
\newblock {{SMOTE: synthetic minority over-sampling technique}}.
\newblock Journal of artificial intelligence research. 2002;16:321--357.

\bibitem{tomek1976two}
Tomek I.
\newblock {{Two modifications of CNN}}. 1976.

\bibitem{tomek1976experiment}
Tomek I.
\newblock {{An experiment with the edited nearest-neighbor rule}}. 1976.

\bibitem{fawcett2006introduction}
Fawcett T.
\newblock {{An introduction to ROC analysis}}.
\newblock Pattern Recognition Letters. 2006;27(8):861--874.

\bibitem{lu2019dying}
Lu L, Shin Y, Su Y, Karniadakis GE.
\newblock {{Dying ReLU and initialization: Theory and numerical examples}}.
\newblock arXiv preprint arXiv:1903.06733; 2019.

\bibitem{kaggle_alzheimers}
Dubey S.
\newblock {{Alzheimer's Dataset (4 class of images)}}. Available from: \url{https://www.kaggle.com/datasets/tourist55/alzheimers-dataset-4-class-of-images}. 2020.





\end{thebibliography}
\end{document}